\documentclass[10pt,journal,compsoc]{IEEEtran}
\ifCLASSOPTIONcompsoc
  \usepackage[nocompress]{cite}
\else
  \usepackage{cite}
\fi
\ifCLASSINFOpdf
   \usepackage[pdftex]{graphicx}
   \graphicspath{{../pdf/}{../jpeg/}{./Figures/}}
   \DeclareGraphicsExtensions{.pdf,.jpeg,.png}
\else
\fi
\usepackage{amsmath}
\usepackage{amssymb}
\usepackage{amsthm}

\usepackage[ruled]{algorithm2e}
\begin{document}
%
\title{Differentiable Projection for Constrained Deep Learning}
%
%
%
%

\author{Dou~Huang,
        Haoran~Zhang,
        Xuan~Song,
        and~Ryosuke~Shibasaki

\IEEEcompsocitemizethanks{\IEEEcompsocthanksitem D. Huang was with the Center for Spatial Information Science, the University of Tokyo.
E-mail: huangd@csis.u-tokyo.ac.jp
\IEEEcompsocthanksitem H. Zhang, X, Song and R. Shibasaki are with the University of Tokyo.}
\thanks{Manuscript received}}

\IEEEtitleabstractindextext{%
\begin{abstract}
Deep neural networks (DNNs) have achieved extraordinary performance in solving different tasks in various fields. However, the conventional DNN model is steadily approaching the ground-truth value through loss backpropagation. In some applications, some prior knowledge could be easily obtained, such as constraints which the ground truth observation follows. Here, we try to give a general approach to incorporate information from these constraints to enhance the performance of the DNNs. Theoretically, we could formulate these kinds of problems as constrained optimization problems that KKT conditions could solve. In this paper, we propose to use a differentiable projection layer in DNN instead of directly solving time-consuming KKT conditions. The proposed projection method is differentiable, and no heavy computation is required. Finally, we also conducted some experiments using a randomly generated synthetic dataset and image segmentation task using the PASCAL VOC dataset to evaluate the performance of the proposed projection method. Experimental results show that the projection method is sufficient and outperforms baseline methods.
\end{abstract}

\begin{IEEEkeywords}
Deep learning, orthogonal projection, constrained learning
\end{IEEEkeywords}}

\maketitle

\IEEEdisplaynontitleabstractindextext

%
\IEEEpeerreviewmaketitle

\IEEEraisesectionheading{\section{Introduction}\label{sec:introduction}}

%
%
%
%
\IEEEPARstart{D}{eep} neural network (DNN) models have been widely used to solve many tasks in different domains and easily outperform many traditional methods. People use the DNN model to replace the existing traditional methods because the DNN models bring better accuracy than traditional methods or because the DNN models save much time for calculations. For example, there are many existing studies devoted to solving various complex optimization problems using DNN models \cite{snoek2015scalable, fischetti2018deep, sun2018learning}. In the field of computer vision or natural language processing, it is more common to use more specific and efficient neural networks such as a convolutional neural network (CNN) \cite{sainath2013deep, xu2014deep, krizhevsky2012imagenet} or recurrent neural network (RNN) \cite{medsker2001recurrent, mikolov2011extensions, schuster1997bidirectional} and its variant models. It has become a consensus for solving various complex problems and has achieved better and more powerful performance than traditional methods.

However, above mentioned models, if trained under the framework of supervised learning or weakly-supervised learning, usually learned how to simulate the ground-truth data through the backpropagation of error between the network output and the ground-truth observation. In this process, the neural network as a "black box" is often criticized that the learning process of the network itself is a process of simulating the truth value, rather than understanding some of the basic rules of the ground-truth value that they should have learned \cite{castelvecchi2016can}. In many implementations, regularization is added to DNN models' loss function to ensure the trained models satisfy some simple constraint conditions. For example, L1-regularization is widely considered to add sparsity to the predictions of trained models. Nevertheless, the problem lies in that what if the ground-truth data set is observed to satisfy some other simple constraint conditions that cannot be regularized by some conventional regularization widely used in loss function? 

\begin{figure}
    \centering
    \includegraphics[width=0.4\textwidth]{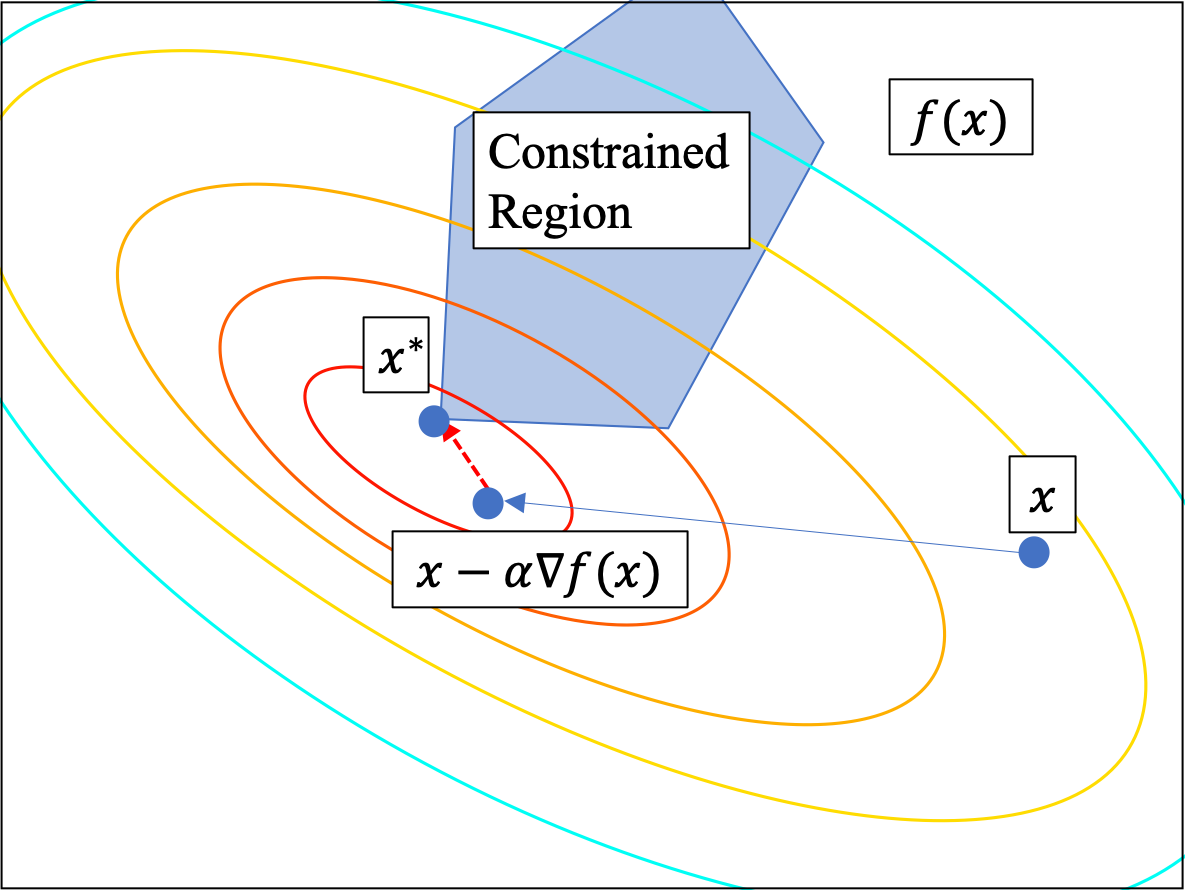}
    \caption{Ground truth observation ($x^*$) belongs to a constrained region formed by some prior knowledge. The blue line indicates the process of optimizing a conventional DNN model by the gradient descent method. The optimized prediction is not satisfied the constraint conditions formed by linear constraint conditions. So the problem is how can we propose a method (red dash line) to obtain the feasible prediction. }
    \label{fig:pd}
\end{figure}

Here we consider some application scenarios such as shown in figure \ref{fig:pd}. We have a ground-truth observation in the constrained region formed by some linear constraints in this scenario. Since it is known that all inference output of DNN models should also belong to the constrained region, we can consider those constraints as a kind of prior knowledge. For a conventional DNN model, the training process directly minimizes the error between ground truth observation and network output. However, an additional process should be optimizing the network output by constrained region, indicated by the red dash line in figure \ref{fig:pd}. By combining these two processes, the model is expected to give some outputs closer to the constrained region, which will make the output closer to the ground truth observation. 


This paper proposes a more unified problem formulation and uses a general framework to solve the DNN models' output that satisfies certain constraints. We hope that the output of the neural network model of supervised or weakly supervised learning is no longer a simple process of simulating the ground-truth value but meets certain constraints. At the same time, we believe that when the output of the DNN models satisfies some of the constraint rules, the interpretability and transferability of the network can be improved. Here, we try to describe this problem in detail as much as possible and show the feasibility of a general method to solve this problem.

\section{Related Works}

Neural network models incorporating constraints are also used in solving optimization problems \cite{xia2004extended, xia2002projection}. There is also some novel neural network model for solving variational inequalities with linear and nonlinear constraints \cite{gao2005novel}. In addition, many people optimize the DNN model by adjusting the internal structure of the neural network to meet certain constraints \cite{han2008modified, chen2013primal}. In recent years, we observed many existing works on constrained DNN models in the field of computer vision, such as a constrained convolutional layer, to accomplish the problem that state-of-the-art approaches cannot detect resampling in re-compressed images initially compressed with high-quality factor \cite{bayar2017robustness}. Besides, a constrained neural network (CCNN) uses a novel loss function to optimize for any set of linear constraints on the output space \cite{pathak2015constrained, bayar2018constrained}. 
Many previous studies on constrained neural network models are mainly dedicated to solving specific problems in one particular field, and there is a lack of a more general description of the problem. Besides, directly combing the DNN models with constrained optimization algorithms, such as the CCNN mentioned above, cannot handle some more generalized constraints and is not time-efficiency since it is necessary to solve optimization problems for every batch network output. 

\section{Deep Neural Networks with constraints}
\subsection{Problem definition}

Given a dataset $D$ that consists of input points $\{\mathbf{x}_i \in \mathbb{R}^n,i=1,2,\cdots,N\}$ and its corresponding ground truth output points $\{\mathbf{y}_i | \mathbf{y}_i \in \mathbb{R}^m, \mathbf{g}_j (\mathbf{y}_i) \leq b_j, i=1, 2, \cdots, N,j=1,2,\cdots,M\}$, in which we assume that ground truth output $\mathbf{y}_i$ satisfy some equality and inequality constraint conditions. Inequality constraints $\mathbf{g}_j (\cdot)$ are usually given by some physical or expert knowledge, which define a feasible set of possible inferenced outputs. In this research, we only focus on the linear constraints problem, then aforementioned constraint conditions can be rewritten as $\{\mathbf{g}_j (\mathbf{y}_i)=\mathbf{a}_j^T \mathbf{y}_i \leq b_j, j=1,2,\cdots,M\}$, where $\mathbf{a}_j \in \mathbb{R}^m, b_j \in \mathbb{R}$. 
Given a deep learning model, denoted by $\mathbf{f}_{\mathbf{\theta}} (\cdot)$, parameterized by $\mathbf{\theta}$, we use $\mathbf{f}_{\mathbf{\theta}} (\mathbf{x}_i)$ denotes the output of this deep learning model. 

For a conventional DNN, the outputs of the model is usually optimized by $\ell (\mathbf{\theta}) + \delta \mathbf{\theta}^T \mathbf{\theta}$, where $\ell(\mathbf{\theta}) = \frac{1}{N} \sum_i {\ell (\mathbf{f}_{\mathbf{\theta}} (\mathbf{x}_i), \mathbf{y}_i})$, and $\delta \mathbf{\theta}^T \mathbf{\theta}$ is a $L_2$ regulariser of weights in the DNN model. Although, this widely-used loss function consists of a regularization for weights in the DNN models. It is designed mainly for optimizing a DNN model which has as simple weights as possible to prevent over-fitting problem, then increase the robustness of the model for approximating some complex mapping from inputs to outputs. However, this objective function used in conventional DNNs lack the ability to ensure a feasible outputs of the model to satisfy the inequality constraints defined in Eq. \ref{eq:obj}. But it is obvious that if the error between inferenced output of the deep learning model and the ground truth is small enough, the inferenced output should eventually satisfy the constraint conditions. Therefore, we consider the minimization of error as the ultimate goal, while the constraint conditions as a kind of prior knowledge, which could help us solve the minimization problem if utilized properly.

Therefore, our objective is to solve the following equation:
\begin{equation} \label{eq:obj}
\begin{aligned}
    & \min_{\mathbf{\theta}} \frac{1}{N} \sum_i {\ell(\mathbf{f}_{\mathbf{\theta}}(\mathbf{x}_i), \mathbf{y}_i)} + \delta \mathbf{\theta}^T \mathbf{\theta} \\
    s.t.& \mathbf{g}_j (\mathbf{f}_{\mathbf{\theta}}(\mathbf{x}_i)) = \mathbf{a}_j^T \mathbf{f}_{\mathbf{\theta}}(\mathbf{x}_i) \leq b_j, j = 1, 2, ..., M 
\end{aligned}
\end{equation}
To train a deep learning model, the most widely used optimization method is probably Gradient Descent (GD) since GD is a standard way to solve unconstrained optimization problem. Therefore, the intuition of solving above objective function is Projected Gradient Descent (PGD), which is considered as a simple way to solve constrained optimization problem. But unfortunately, PGD is not considered as an efficient method to solve optimization problem consists of some complex constraint conditions. Therefore, we aim to propose a more generalized efficient optimization method for above objective function. 
\subsection{Preliminary}
\subsubsection{Projected Gradient Descent (PGD)}
Let us start with a simple unconstrained optimization problem like: $\min_{\mathbf{x} \in \mathbb{R}^n} f (\mathbf{x})$. The optimal solution $\mathbf{x}^*$ can be found using GP by: $\mathbf{x}_{t+1} \leftarrow \mathbf{x}_t - \alpha \nabla f (\mathbf{x}_t)$, where $\alpha > 0$ is the step size, and $\nabla f$ is the gradient of $f$. However, if $\mathbf{x}$ is defined in a constrained region, denoted as $\mathcal{Q} \subset \mathbb{R}^n$, then the problem becomes $\min_{\mathbf{x} \in \mathcal{Q}} f(\mathbf{x})$. The GD fails for this problem because $\mathbf{x}_t - \alpha \nabla f(\mathbf{x}_t)$ is not guaranteed to be inside the constrained region. An intuition to solve this problem is using projection $P_\mathcal{Q} (\mathbf{x}^*) \equiv \arg\min_{\mathbf{x} \in \mathcal{Q}} \frac{1}{2} ||\mathbf{x} - \mathbf{x}^*||^2$ to find the closest point in the constrained region of updated $\mathbf{x}_{t+1}$ by $\mathbf{x}_{t+1} \leftarrow P_{\mathcal{Q}} (\mathbf{x}_t - \alpha \nabla f(\mathbf{x}_t))$. 

The above projection is defined as an implicit function, and this projection function itself defines a constrained optimization problem. In this paper, we consider constrained problems only contain linear constraint conditions, so we can define the constrained region as $\mathcal{Q} = \{\mathbf{x} | \mathbf{g} (\mathbf{x}) \leq \mathbf{b}\}$, then we can rewrite the projection function as $\min_{\mathbf{x}} \frac{1}{2} ||\mathbf{x} - \mathbf{x}^*||^2, s.t. \mathbf{g} (\mathbf{x}) \leq \mathbf{b}$. To solve this equation, we could use the Karush–Kuhn–Tucker (KKT) conditions: 
\begin{equation} \label{eq:kkt}
\begin{aligned}
& \frac{\partial}{\partial \mathbf{x}} \frac{1}{2} ||\mathbf{x} - \mathbf{x}^*||^2 + \sum_{j=1}^M {\lambda_j \frac{\partial}{\partial \mathbf{x}} \mathbf{g}_j (\mathbf{x})} = 0, \\
& \mathbf{g}_j (\mathbf{x}) - b_j \leq 0, \lambda_j (\mathbf{g}_j (\mathbf{x}) - b_k) = 0, \\
& \lambda_j \geq 0, j = 1, 2,\cdots, M
\end{aligned}
\end{equation}
Where $\lambda_j$ is constant value. The above Eq. \ref{eq:kkt} gives an analytic solution of the constrained optimization problem. 

Although the above Eq. \ref{eq:kkt} is usually not solved directly, many optimization algorithms can be interpreted as methods for numerically solving this KKT conditions \cite{boyd2004convex}. Therefore, PGD for constrained optimization problem can be written as a two-step update process: 1) $\mathbf{x}_{t+\frac{1}{2}} \leftarrow \mathbf{x}_t - \alpha \nabla f(\mathbf{x}_t)$; 2) solving KKT conditions of constrained optimization problem $\mathbf{x}_{t+1} \leftarrow \arg\min_{\mathbf{x} \in \mathcal{Q}} \frac{1}{2} ||\mathbf{x} - \mathbf{x}_{t + \frac{1}{2}}||^2$. Both of these two steps can be numerically solved according to aforementioned method. 

\subsubsection{PGD for optimizing DNNs}
The optimizing process of a conventional DNN model can be written as following equation: 
\begin{equation} \label{eq:rmsprop}
\begin{aligned}
    \mathbf{\theta}_{t+1} & \leftarrow \mathbf{\theta}_t - \alpha \frac{\frac{1}{M} \sum_{i \in \mathcal{M}_t} \nabla_{\mathbf{\theta}_t} \ell (\mathbf{f}_{\mathbf{\theta}_t} (\mathbf{x}_i), \mathbf{y}_i) + \delta \mathbf{\theta}_t}{\sqrt{s_{t+1}} + \epsilon} \\
    \mathbf{s}_{t+1} & \leftarrow (1 - \beta) \mathbf{s}_t - \beta (\frac{1}{M} \sum_{i \in \mathcal{M}_t} \nabla_{\mathbf{\theta}_t} \ell (\mathbf{f}_{\mathbf{\theta}_t} (\mathbf{x}_i), \mathbf{y}_i) + \delta \mathbf{\theta}_t)^2
\end{aligned}
\end{equation}
Where $\alpha > 0$ and $0 < \beta < 1$ are learning rates, $t$ is the iteration, $\epsilon$ is a small scalar, and $\mathcal{M}_t$ is a mini batch contains $M$ data samples. This simple update scales well and achieves good performance even in some very large problems. 

Here we only consider the case that the output of the deep learning model $\mathbf{f}_{\mathbf{\theta}} (\mathbf{x}_i)$ should satisfy the linear constraint conditions, and we define a implicit projection function denoted as $P$. From the definition of the point $\mathbf{y}_i$, it is obvious that all $\mathbf{y}_i$ should located in the feasible region of the linear constraint condition. At that same time, since an optimal $P(\mathbf{f}_{\mathbf{\theta}} (\mathbf{x}_i))$ should be the closest with $\mathbf{f}_{\mathbf{\theta}} (\mathbf{x}_i)$, we know that $P(\mathbf{f}_{\mathbf{\theta}} (\mathbf{x}_i))$ should be on the boundary of the feasible region, and the line connected by $P(\mathbf{f}_{\mathbf{\theta}} (\mathbf{x}_i))$ and $\mathbf{f}_{\mathbf{\theta}} (\mathbf{x}_i)$ is orthogonal with the hyperplane where $P(\mathbf{f}_{\mathbf{\theta}} (\mathbf{x}_i))$ located on since $\frac{\partial}{\partial \mathbf{f}_{\mathbf{\theta}} (\mathbf{x}_i)} \mathcal{L} (\mathbf{f}_{\mathbf{\theta}} (\mathbf{x}_i), \lambda) = 0$. In the case that all constraint will not be active when the output $\mathbf{f}_{\mathbf{\theta}} (\mathbf{x}_i)$ satisfy the constraint conditions, then the projected point $P(\mathbf{f}_{\mathbf{\theta}} (\mathbf{x}_i))$ is the same with the output $\mathbf{f}_{\mathbf{\theta}} (\mathbf{x}_i)$ exactly. However, if the output of DNN model is not satisfy the constraint conditions, we should add this projection function $P$ in the optimizing process of DNN model. To revise update process for DNN model weights $\mathbf{\theta}$, we should replace $\mathbf{f}_{\mathbf{\theta}_t} (\mathbf{x}_i)$ with $P(\mathbf{f}_{\mathbf{\theta}_t} (\mathbf{x}_i))$ in Eq. \ref{eq:rmsprop}. We know that although we could solve KKT conditions to get the projected output, but usually it can only be obtained through some numerical constrained optimization methods. So to optimize weights in DNN models, we need firstly optimizing the projected output at each iteration. 

Although we don't know the explicit function $P$, we can obtain an identical point $P(\mathbf{f}_{\mathbf{\theta}} (\mathbf{x}_i))$ by solving aforementioned Eq. \ref{eq:kkt}, then we are able to find the optimal results, which is the closest points of original network outputs in the constrained region. However, directly solving this constrained optimization problem is not an efficient solution since it is computed every iteration when the DNN produces a mini batch of outputs. We consider this training process as a two-step update method similar with the PGD method. For example, constrained convolutional neural network \cite{pathak2015constrained} is proposed to solve image segmentation with some simpleconstraint conditions, which is required to solves a constrained optimization problem every time at each iteration when training the model. We doubt this strategy could be used in more generalized scenarios. Actually, we also concern that finding the closet point $\mathbf{f}_{\mathbf{\theta}} (\mathbf{x}_i)$ of original DNN output in the constrained region is not necessary. That is because what we want to achieve is that the output of DNN should be close to the constrained region. In the best situation, the output of DNN should satisfy the constraint. So directly solving KKT conditions for guiding the training process of DNN model is waste of computational resource. Therefore, we aim to find other approach to solve the Eq. \ref{eq:obj}. 

\subsection{Defferentiable Projection DNN}
\subsubsection{Projection for linear equality constraint conditions}

Let us consider a specialized scenario that there is only linear equality constraint conditions for the output of deep learning models. We can write the constrained region into matrix form like $\mathbf{a}^T \mathbf{f}_{\mathbf{\theta}} (\mathbf{x}_i) = \mathbf{b}$, where $\mathbf{a} \in \mathbb{R}^{m, M}, \mathbf{b} \in \mathbb{R}^M$. We can easily obtain the condition for $\mathbf{f}_{\mathbf{\theta}} (\mathbf{x}_i)$ have feasible solutions is that $rank(\mathbf{a}) \leq M \leq m$. It means that there exists a feasible area when the number of efficient constraints $\mathbf{a}_j$ is less than the dimension of the variable $\mathbf{f}_{\mathbf{\theta}} (\mathbf{x}_i)$. 
More specially, if the rank of $\mathbf{a}$ satisfy that $rank(\mathbf{a}) = M$, we know that given any $\mathbf{y}_i \in \mathbb{R}^m$, the closest point $\mathbf{y}_i^*$ subject to $\mathbf{a}^T \mathbf{y}_i = \mathbf{b}$ is $\mathbf{y}_i^* = (\mathbf{I} - \mathbf{a} (\mathbf{a}^T \mathbf{a})^{-1} \mathbf{a}^T) \mathbf{y}_i + \mathbf{a} (\mathbf{a}^T \mathbf{a})^{-1} \mathbf{b}$. 

The proof of above equation is given in Appendix. Based on this equation, given any network output $\mathbf{f}_{\mathbf{\theta}} (\mathbf{x}_i)$, we can find the closest point. At first, let us simplify the notation using  $P(\mathbf{f}_{\mathbf{\theta}} (\mathbf{x}_i)) = (\mathbf{I} - \mathbf{a} (\mathbf{a}^T \mathbf{a})^{-1} \mathbf{a}^T) \mathbf{f}_{\theta} (\mathbf{x}_i) + \mathbf{a} (\mathbf{a}^T \mathbf{a})^{-1} \mathbf{b}$. We should notice that $\ell(P(\mathbf{f}_{\mathbf{\theta}} (\mathbf{x}_i)), \mathbf{y}_i) \leq \ell(\mathbf{f}_{\mathbf{\theta}} (\mathbf{x}_i), \mathbf{y}_i)$ if $\ell$ is defined as Euclidean distance because 
\begin{equation} \label{eq:l2}
\begin{aligned}
  \ell(&\mathbf{f}_{\mathbf{\theta}} (\mathbf{x}_i), \mathbf{y}_i) = ||\mathbf{f}_{\mathbf{\theta}} (\mathbf{x}_i) - \mathbf{y}_i||_2^2 \\
  & = ||P(\mathbf{f}_{\mathbf{\theta}} (\mathbf{x}_i)) - \mathbf{y}_i + \mathbf{f}_{\mathbf{\theta}} (\mathbf{x}_i) - P(\mathbf{f}_{\mathbf{\theta}} (\mathbf{x}_i))||_2^2 \\ 
  & = ||P(\mathbf{f}_{\mathbf{\theta}} (\mathbf{x}_i)) - \mathbf{y}_i||_2^2 + ||\mathbf{f}_{\mathbf{\theta}} (\mathbf{x}_i) - P(\mathbf{f}_{\mathbf{\theta}} (\mathbf{x}_i))||_2^2 \\ & + 2 (\mathbf{f}_{\mathbf{\theta}} (\mathbf{x}_i) - P(\mathbf{f}_{\mathbf{\theta}} (\mathbf{x}_i)))^T (P(\mathbf{f}_{\mathbf{\theta}} (\mathbf{x}_i)) - \mathbf{y}_i) \\
  & \geq ||P(\mathbf{f}_{\mathbf{\theta}} (\mathbf{x}_i)) - \mathbf{y}_i||_2^2 = \ell(P(\mathbf{f}_{\mathbf{\theta}} (\mathbf{x}_i)), \mathbf{y}_i)
\end{aligned}
\end{equation}
Where $(\mathbf{f}_{\mathbf{\theta}} (\mathbf{x}_i) - P(\mathbf{f}_{\mathbf{\theta}} (\mathbf{x}_i)))^T (P(\mathbf{f}_{\mathbf{\theta}} (\mathbf{x}_i)) - \mathbf{y}_i) = 0$, and $||\mathbf{f}_{\mathbf{\theta}} (\mathbf{x}_i) - P(\mathbf{f}_{\mathbf{\theta}} (\mathbf{x}_i))||_2^2 \geq 0$. Therefore, we can achieve a conclusion that the projected output $P(\mathbf{f}_{\mathbf{\theta}} (\mathbf{x}_i))$ is a better inference because that the distance between projected output with ground truth is smaller than the original output. 

There are two strategy of using the property of projection for linear equality constraint conditions in Eq. \ref{eq:l2}. 1) We train a conventional DNN model at first, and then we can add a linear projection layer behind the pre-trained DNN model to obtain the projected outputs. 2) We add a linear projection layer behind the conventional DNN model, and then train the model using loss calculated by projected inference and ground truth. 

For the first strategy, we notice that the DNN model can produce better inference which is closer with the ground truth even with fine-tuning the parameters of the pre-trained DNN model based on the above equations. This indicates that if we know some very basic prior knowledge about the ground truth data, we can build some linear equality constraint conditions and the corresponding projection layer. Using this simple projection layer, we can improve the performance of pre-trained models without fine-tuning their parameters, which is very time-efficient and require less computation resource. 

For the second strategy, here we analyze the change of training process at first. According to the chain rule used in Back Propagation (BP) algorithm, we can look at the final layer at the beginning. In above context, we use $\mathbf{f}_{\mathbf{\theta}} (\mathbf{x}_i)$ denote the final output of the model, but since here we only consider the final layer, we use the $z^L$ denote the output of final layer $L$, and then $\mathbf{f}_{\mathbf{\theta}} (\mathbf{x}_i) = \sigma (z^L)$ if we take sigmoid activation function as an example. For a conventional DNN model, the local error of the final layer can be calculated by $\delta^L = \nabla_{\mathbf{f}_{\mathbf{\theta}} (\mathbf{x}_i)} \ell \odot \sigma^{'} (z^L)$. Then, we can calculate the local error of previous layers, which is used to calculate the local gradient of parameters in each layers using BP algorithm. However, if we add a projection layer behind the conventional DNN model, even we didn't change the structure of the model, the local gradient of parameters will be effected because the local error of the final layer changed to $\delta^L = \nabla_{P(\mathbf{f}_{\mathbf{\theta}} (\mathbf{x}_i))} \ell \odot P \sigma^{'} (z^L)$. This will not increase the computation burden, and we can regard this as a special activation function using deep learning concept.

\subsubsection{Projection for linear inequality constraint conditions}

While for a more generalized scenario, the constraints are not always independent with each other, but also there could exist some inequality constraint conditions in many applications. This section, we will give a projection method for linear inequality constraint problems. Without any assumption about linear independent of constraints in Eq. \ref{eq:obj}, 
we have following definitions. At first, we need define a index set of violated constraint conditions, denoted by $I(\mathbf{f}_{\theta}(\mathbf{x}_i)) = \{j| \mathbf{a}_j^T \mathbf{f}_{\theta}(\mathbf{x}_i) - b_j > 0\}$. 
Then, we can define a iterative projection like: 
\begin{equation} \label{eq:projection}
    \begin{aligned}
      \mathbf{y}_i^{t+1} = \mathbf{y}_i^t - \lambda \sum_{j \in I(\mathbf{y}_i^t)} \pi_j^t \frac{\mathbf{a}_j (\mathbf{a}_{j}^T \mathbf{y}_i^t - b_{j})}{\mathbf{a}_{j}^T \mathbf{a}_{j}}
    \end{aligned}
\end{equation}
where $0<\lambda<2$, and $\pi_j^t$ is weights of selected constraint conditions. The value of the weights can be determined once $\pi_j^t > \gamma$, where $\gamma$ is a small positive quantity because we need make sure every violated constraint conditions is considered. We can use $\pi^t = 1 / |I(\mathbf{y}_i^t)|$ as a normalized weights of violated constraint conditions for simplification. 
With above definition, we can use Eq. \ref{eq:projection} to find a feasible solution, which satisfy constraint conditions of Eq. \ref{eq:obj}. Given any output $\mathbf{y}_i^0 = \mathbf{f}_{\mathbf{\theta}} (\mathbf{x}_i)$ of DNN models, we know that $\mathbf{y}_i^{t} \in \mathcal{Q}, t \to \infty$,
where $\mathcal{Q}$ is the feasible region defined by constraint conditions in Eq. \ref{eq:obj}. 
The detailed proof of the convergence of above algorithm is given in Appendix. 

So far, we propose to build a projection method like Eq. \ref{eq:projection} to find a feasible solution in constrained region when given any output of DNN models after many iterations. But how many iterations are needed to find such feasible solution is an issue to be solved. At the same time, we also concern that many projection iterations will lead to the increasing of computation time, which will make the projection method unsuitable for practical implementation. However, in previous section, we notice that there is a good property that for projected output is closer to the feasible region after each projection in linear equality case, so we are curious about that whether this property exists in linear inequality case. 

At first, we introduce a surrogate constraint \cite{yang1992new}, denoted by $\pi^T \mathbf{a}_{I(\mathbf{y}_i)}^T \mathbf{y}_i = \pi^T \mathbf{b}_{I(\mathbf{y}_i)}$, where $\pi, \mathbf{b}_{I(\mathbf{y}_i)} \in \mathbb{R}^{|I(\mathbf{y}_i)|}$, and $\mathbf{a}_{I(\mathbf{y}_i)} \in \mathbb{R}^{m, |I(\mathbf{y}_i)|}$. Then, we can rewrite previous Eq. \ref{eq:projection} like: 
\begin{equation} \label{eq:proj_matrix}
    \mathbf{y}_i^{t+1} = \mathbf{y}_i^t - \frac{\lambda (\pi^T \mathbf{a}_{I(\mathbf{y}_i^t)}^T \mathbf{y}_i^t - \pi^T \mathbf{b}_{I(\mathbf{y}_i^t)})}{||\mathbf{a}_{I(\mathbf{y}_i^t)} \pi||_2^2} \mathbf{a}_{I(\mathbf{y}_i^t)} \pi
\end{equation}
let us define an error vector between output after $t$ steps projection and ground truth $\mathbf{y}_i$ , denoted as $\mathbf{e}_i^t = \mathbf{y}_i^t - \mathbf{y}_i$. Then, based on previous equation, we have following equation: 
\begin{equation} \label{eq:derivation}
    \begin{aligned}
      \mathbf{e}_i^{t+1} &= \mathbf{e}_i^t +\mathbf{y}_i^{t+1} - \mathbf{y}_i^t \\
      &= \mathbf{e}_i^t - \frac{\lambda (\pi^T \mathbf{a}_{I(\mathbf{y}_i^t)}^T \mathbf{y}_i^t - \pi^T \mathbf{b}_{I(\mathbf{y}_i^t)})}{||\mathbf{a}_{I(\mathbf{y}_i^t)} \pi||_2^2} \mathbf{a}_{I(\mathbf{y}_i^t)} \pi \\
    \end{aligned}
\end{equation}
Based on above Eq. \ref{eq:derivation}, we can write following equation:
\begin{equation} \label{eq:l2norm}
    \begin{aligned}
      &||\mathbf{e}_i^{t+1}||_2^2 = ||\mathbf{e}_i^t||_2^2 + \lambda^2 \frac{(\pi^T \mathbf{a}_{I(\mathbf{y}_i^t)}^T \mathbf{y}_i^t - \pi^T \mathbf{b}_{I(\mathbf{y}_i^t)})^2}{||\mathbf{a}_{I(\mathbf{y}_i^t)} \pi||_2^2} \\
      &- 2 \lambda \frac{\pi^T \mathbf{a}_{I(\mathbf{y}_i^t)}^T \mathbf{y}_i^t - \pi^T \mathbf{b}_{I(\mathbf{y}_i^t)}}{||\mathbf{a}_{I(\mathbf{y}_i^t)} \pi||_2^2} (\mathbf{a}_{I(\mathbf{y}_i^t)} \pi)^T (\mathbf{y}_i^t - \mathbf{y}_i) \\
      &= ||\mathbf{e}_i^t||_2^2 - \lambda (2 - \lambda) \frac{(\pi^T \mathbf{a}_{I(\mathbf{y}_i^t)}^T \mathbf{y}_i^t - \pi^T \mathbf{b}_{I(\mathbf{y}_i^t)})^2}{||\mathbf{a}_{I(\mathbf{y}_i^t)} \pi||_2^2} \\
      &+ 2 \lambda \frac{(\pi^T \mathbf{a}_{I(\mathbf{y}_i^t)}^T \mathbf{y}_i^t - \pi^T \mathbf{b}_{I(\mathbf{y}_i^t)}) (\pi^T \mathbf{a}_{I(\mathbf{y}_i^t)}^T \mathbf{y}_i - \pi^T \mathbf{b}_{I(\mathbf{y}_i^t)})}{||\mathbf{a}_{I(\mathbf{y}_i^t)} \pi ||_2^2} 
    \end{aligned}
\end{equation}
From above equation, we know that $||\mathbf{e}_j^{t+1}||_2^2 \leq ||\mathbf{e}_j^t||_2^2$ because the second and third term in Eq. \ref{eq:l2norm} is non-positive. 
We know that we could eventually find a projected output $\mathbf{y}^{t+1}_i$ in feasible region given the deep leaning model output $\mathbf{f}_{\mathbf{\theta}} (\mathbf{x}_i)$, but this process could be time-consuming potentially since many projection steps is necessary. But fortunately, there is a good property of this projection method given by above derivation, which indicates that projected points at current step are always closer to the constrained region than those projected points at previous step. 

Therefore, only limited steps of projection is needed, and the projected outputs are always better than original deep learning models' outputs. 
Here we make small revision for the definition of weights for selected violated constraint conditions as $\pi = \frac{1}{M} \frac{\partial}{\partial \mathbf{l}^t} \max (\mathbf{l}^t, \mathbf{0})$, where $\mathbf{l}^t = \mathbf{a}^T \mathbf{y}_i^t - \mathbf{b}$. 
We can simplify Eq. \ref{eq:proj_matrix} as: 
\begin{equation} \label{eq:proj_simplified}
    \mathbf{y}_i^{t+1} = \mathbf{y}_i^t - \frac{\lambda (\pi^T \mathbf{a}^T \mathbf{y}_i^t - \pi^T \mathbf{b})}{||\mathbf{a} \pi||_2^2 + \epsilon} \mathbf{a} \pi
\end{equation}
Where $\epsilon$ is a very small scalar because we notice that $\pi = \mathbf{0}$ when $\mathbf{y}_i^t \in \mathcal{Q}$. When the set of violated constraint conditions is not empty, the very small scalar $\epsilon$ has little effect on the positive value of $||\mathbf{a} \pi||_2^2$. However, we need $\epsilon$ to prevent the denominator of the equation to be $0$ when $\pi = \mathbf{0}$. 
We notice that there is no assumption about the linear independent of constraint conditions, which means that the rank of $\mathbf{a}$ is not always to be full row rank, so we summarize the above equations in a more uniform differentiable way.

\begin{algorithm}
\caption{Differentiable projection deep learning}
\renewcommand{\KwData}{\textbf{Input: }}
\renewcommand{\KwResult}{\textbf{Output: }}
\KwData {$\mathbf{x}_i \in \mathbb{R}^n, i=1,2,...,N$} \;
\KwResult {$\mathbf{y}_i \in \mathbb{R}^m, s.t. \mathbf{a}_j^T \mathbf{y}_i \leq b_j, j=1,2,...,M$} \;
\textbf{Initialization}: $\mathbf{f}_{\theta}(\cdot), epochs, layers, \alpha, \beta, \epsilon, \mathbf{a}, \mathbf{b}, \lambda$ \;
$k \gets 0$ \;
\While{$k \leq epochs$}{
$t \gets 0$ \;
$\mathbf{y}_i^t \gets \mathbf{f}_{\theta_{k}}(\mathbf{x}_i)$ \;
\While{$t \leq layers$}{
$\mathbf{l}^t \gets \mathbf{a}^T \mathbf{y}^t - \mathbf{b}$ \;
$\pi \gets \frac{1}{M} \frac{\partial}{\partial \mathbf{l}^t} \max(\mathbf{l}^t, 0)$ \;
$\mathbf{y}_i^t \gets \mathbf{y}_i^t - \frac{\lambda (\pi^T \mathbf{a}^T \mathbf{y}_i^t - \pi^T \mathbf{b})}{||\mathbf{a} \pi||_2^2 + \epsilon} \mathbf{a} \pi$ \;
$t \gets t + 1$ \;
}
$\mathbf{s}_k \gets (1 - \beta) \mathbf{s}_k - \beta (\nabla_{\mathbf{\theta}_k} \ell (\mathbf{f}_{\mathbf{\theta}_k} (\mathbf{x}_i), \mathbf{y}_i^t) + \delta \mathbf{\theta}_k)^2$ \;
$\mathbf{\theta}_k \leftarrow \mathbf{\theta}_k - \alpha \frac{\nabla_{\mathbf{\theta}_k} \ell (\mathbf{f}_{\mathbf{\theta}_k} (\mathbf{x}_i), \mathbf{y}_i^t) + \delta \mathbf{\theta}_k}{\sqrt{s_k} + \epsilon}$ \;
}
\textbf{Return: } $\mathbf{y}_i^t, \mathbf{f}_{\theta_k}(\cdot)$ \;
\end{algorithm}

At each time, the DNN model output some original points $\mathbf{f}_{\theta}(\mathbf{x}_i)$, by using projection method defined in Algorithm 1, there will be some converged feasible points $\mathbf{y}_i^t$ located in constrained region. We have to clarify that the projected feasible points $\mathbf{y}_i^t$ is not always the KKT points of original output, which means that projected feasible points are not always the closest points of original output. By using projection method, we can find some possible feasible solutions but not the best solution. 

\section{Experiments}

\subsection{Numerical Experiments}
In this section, we present some numerical experiments to demonstrate the effectiveness of the constrained neural network. The data we used for numerical experiments are randomly generated synthetic data. In this experiment, we use the same parameter settings and training process for a fair comparison with different methods. The input data is generated by a 64-dimensional unit Gaussian distribution. Then we build a nonlinear function with random parameters to convert the 64-dimensional inputs to 8-dimensional vectors. The linear equality constraints and inequality constraints are all generated randomly, and we only keep the data pair that satisfy these random constraints. The objective of this problem is to evaluate the performance of projection DNN using MSE as the metric. 

\subsubsection{Effectiveness of projection layer}
In this experiment, we compare the projection DNN (PDNN) with a conventional DNN. We also compare the results of conducting projection as post-processing of DNN, denoted as DNN+projection. It can be found that the PDNN gives the best results, while DNN+projection also outperforms the conventional DNN. That proves our assumption that by incorporating some linear constraints as prior information, even a simple model could give better results.  
\begin{table}
    \centering
    \caption{Comparison of results.}
    \begin{tabular}{c c c c}
    \hline
     & DNN & DNN+Proj & PDNN  \\
    \hline
    MSE & 0.02279 & 0.02268 & \textbf{0.02093} \\
    \hline
    \end{tabular}
    \label{tab:comparison}
\end{table}
\subsubsection{ablation}
\begin{figure}[h]
    \centering
    \includegraphics[width=0.45\textwidth]{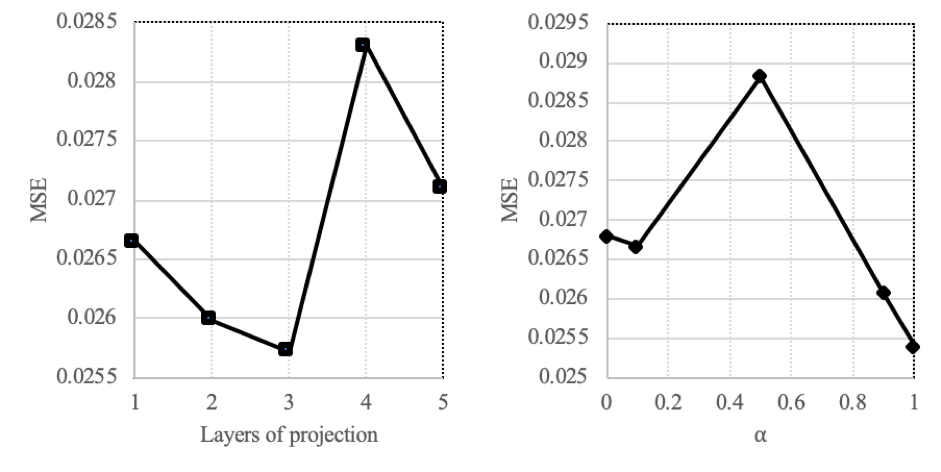}
    \caption{the performance when using different number of projection layer (left); the impact of the hyperparameter $\alpha$ in the loss function (right).}
    \label{fig:ablation}
\end{figure}

We also evaluate the impact of the projection layer and the hyperparameter $\alpha$ in the loss function. For a single iteration, we know that with more projection layers in a PDNN model, the projected outputs will be closer to the ground truth observations. This property will lead to an intuition that we should use more projection layers in a PDNN to get better performance. At the same time, the experimental results deny this kind of intuition shown in Figure \ref{fig:ablation} (left). At least in this experiment, the best performance is obtained when the PDNN consists of 3 stacked projection layers. The number of projection layers in a PDNN to achieve the best performance could differ in other application scenarios. Still, based on these results, it suggests that using limited stack projection layers could be sufficient instead of stacking as many projections as possible, which will lead to an increased computational burden. 
Since the loss function is originally defined as $loss = (1 - \alpha) (\mathbf{y} - \mathbf{y}^t)^2 + \alpha (\mathbf{y} - \mathbf{f}_{\theta}(\mathbf{x}))^2$, we also evaluate the impact of the hyperparameter $\alpha$. The best performance is obtained when $\alpha = 1$ shown in Fig. \ref{fig:ablation} (right). This result suggests that we should only use projected outputs for calculating the loss to guide the training of the PDNN. However, we think it depends on the different scenarios to decide the hyperparameter $\alpha$. 

\subsubsection{Impact of constraints}
Experiments are also conducted for the comparison of different methods to evaluate the performance under different constraints conditions. The loss function of "DNN+resid" in table \ref{tab:constraints} is $loss = (\mathbf{y}^t - \mathbf{y})^2 + max(\mathbf{a}^T \mathbf{y}^t, 0)$. When the network output $\mathbf{y}^t$ did not satisfy the constraints, $\mathbf{a}^T \mathbf{y}^t$ will be a positive value. If the network output satisfies the constraints, then the residual will be 0. The loss function of "DNN+fix resid" is similar. When the network outputs did not satisfy the constraints, give a fixed penalty in the loss function. 
\begin{table}
    \centering
    \caption{The performance of different methods under different constraints conditions}
    \resizebox{0.45\textwidth}{!}{
    \begin{tabular}{c c c c c c}
        \hline
        Constraints & DNN & fixed penalty & penalty & DNN+proj & DPDNN  \\
        \hline
        3*8 & 0.08979 & 0.05991 & 0.06106 & 0.08979 & 0.09090 \\
        \hline
        4*8 & 0.09952 & 0.04378 & 0.04472 & 0.09951 & 0.09872 \\
        \hline
        5*8 & 0.12691 & 0.04946 & 0.05587 & 0.09055 & 0.07386 \\
        \hline
        6*8 & 0.18655 & 0.26512 & 0.26483 & 0.18655 & 0.18650 \\
        \hline
        7*8 & 0.10872 & 0.13634 & 0.13788 & 0.10871 & 0.10817 \\
        \hline
    \end{tabular}}
    \label{tab:constraints}
\end{table}
From table \ref{tab:constraints}, we notice that in most cases, the PDNN outperforms conventional DNN and DNN+proj. We must also figure out that the baseline methods “DNN+resid” and “DNN+resid” sometimes give significantly better results than the other three methods, while they are not robust. When the constraint conditions change, the performance of these two methods becomes worse than the other three methods. The possible reason for this phenomenon could be that when the complexity of the constraint condition increases, the residual of the outputs will increase dramatically, and the model cannot give accurate predictions. This might be solved by tuning the hyperparameters in the model, but it requires more attempt efforts than PDNN or conventional DNN, so another advantage of the PDNN is that we could always get better results without additional hyperparameter tuning efforts. 

\subsection{Image segmentation using weak labels}
We analyze and compare the performance of our proposed differentiable projection layer for image-level tags and some additional supervision such as object size information. The objective is to learn models to predict dense multi-class semantic segmentation for a new image. At first, we train a VGG with pre-trained parameters, and then we add differentiable projection layers behind the model to demonstrate the utility of the proposed projection layer and how it helps increase supervision levels. 

\subsubsection{Dataset}
We evaluate the proposed DPDNN for the task of semantic image segmentation on the PASCAL VOC dataset \cite{everingham2015pascal}. The dataset contains pixel-level labels for 20 object classes and a separate background class. For a fair comparison to prior work, we use a similar setup to train all models. Training is performed only on the VOC 2012 train set. The VGG network architecture used in our algorithm was pre-trained on the ILSVRC dataset for the classification task of 1K classes. Results are reported in the form of standard intersection over union (IoU) metric. It is defined per class as the percentage of pixels predicted correctly out of total pixels labeled or classified as that class. 

\subsubsection{Implementation details}
In this section, we discuss the overall pipeline of our algorithm applied for semantic image segmentation. We consider the weakly supervised setting, i.e., only image-level labels are present during training. At the test time, the task is to predict the semantic segmentation mask for a given image. 
The CNN architecture used in our experiments is derived from the VGG 16-layer network \cite{simonyan2014very}. It was pre-trained on Imagenet 1K class dataset and achieved winning performance on ILSVRC14. We cast the fully connected layers into convolutions in a similar way as suggested in \cite{long2015fully}, and the last fc8 layer with 1K outputs is replaced by that containing 21 outputs corresponding to 20 objects classes in Pascal VOC and background class. The overall network stride of this fully convolutional network is 32s. Also, we do not learn any weights of the last layer from Imagenet, which is unlike \cite{pathak2014fully, pinheiro2015weakly}. Apart from the initial pre-training, all parameters are finetuned only on the Pascal VOC dataset. 

The FCN takes in arbitrarily sized images and produces coarse heatmaps corresponding to each class in the dataset. Although it is not time-consuming for our proposed DP layer to be applied to finer-grained heatmaps, we apply our DP layer to these coarse heatmaps just for a fair comparison with \cite{pathak2015constrained}. The network is trained using SGD with momentum. We follow previous works and train our models with a batch size of 1, momentum of 0.99, and an initial learning rate of $1e-6$. Unlike previous works, we did not apply a fully connected conditional random field model \cite{krahenbuhl2011efficient}.

\subsubsection{constraint conditions}
We use similar constraint conditions used in \cite{pathak2015constrained}. For each training image $I$, we are given a set of image-level labels $\mathcal{L}_I$. 

Suppression constraint: No pixel should be labeled as classes that are not listed in the image-level labels.
\begin{equation} \label{eq:suppression}
    \sum_{i \in I} p_i(l) \leq 0, \forall l \notin \mathcal{L}_I
\end{equation}

Foreground constraint: The number of pixels in the image labeled as classes in the image-level label should be greater than $a_l = 0.05 |I|$. This constraint is used to make sure each label can be detected efficiently. 
\begin{equation} \label{eq:foreground}
    \sum_{i \in I} p_i(l) \geq a_l, \forall l \in \mathcal{L}_I
\end{equation}

Background constraint: Here we set $l=0$ as the background label. The upper bound and lower bound of the number of pixels to be labeled as the background is $a_0=0.3|I|, b_0=0.7|I|$, respectively.
\begin{equation} \label{eq:background}
    a_0 \leq \sum_{i \in I} p_i(0) \leq b_0
\end{equation}
In practical implementation, all constraint conditions listed above are formed as a matrix and can efficiently be solved using a differentiable projection layer. 

\subsubsection{Results}
We summarize the results of experiments in table \ref{tab:my_label}. At first, we train a VGG model, and then we use a different number of differentiable projection layers after the VGG model without additional parameters tuning. From the results in the table, we notice that VGG obtains the best mean IoU with five differentiable projection layers. However, if we look at the performance in a different class, we find that VGG with 20 differentiable projection layers gives the best performance in most classes. Again, the results in this table show that more projection layers cannot guarantee better overall results in practical applications.

\begin{figure*}[h]
    \centering
    \includegraphics[width=\textwidth]{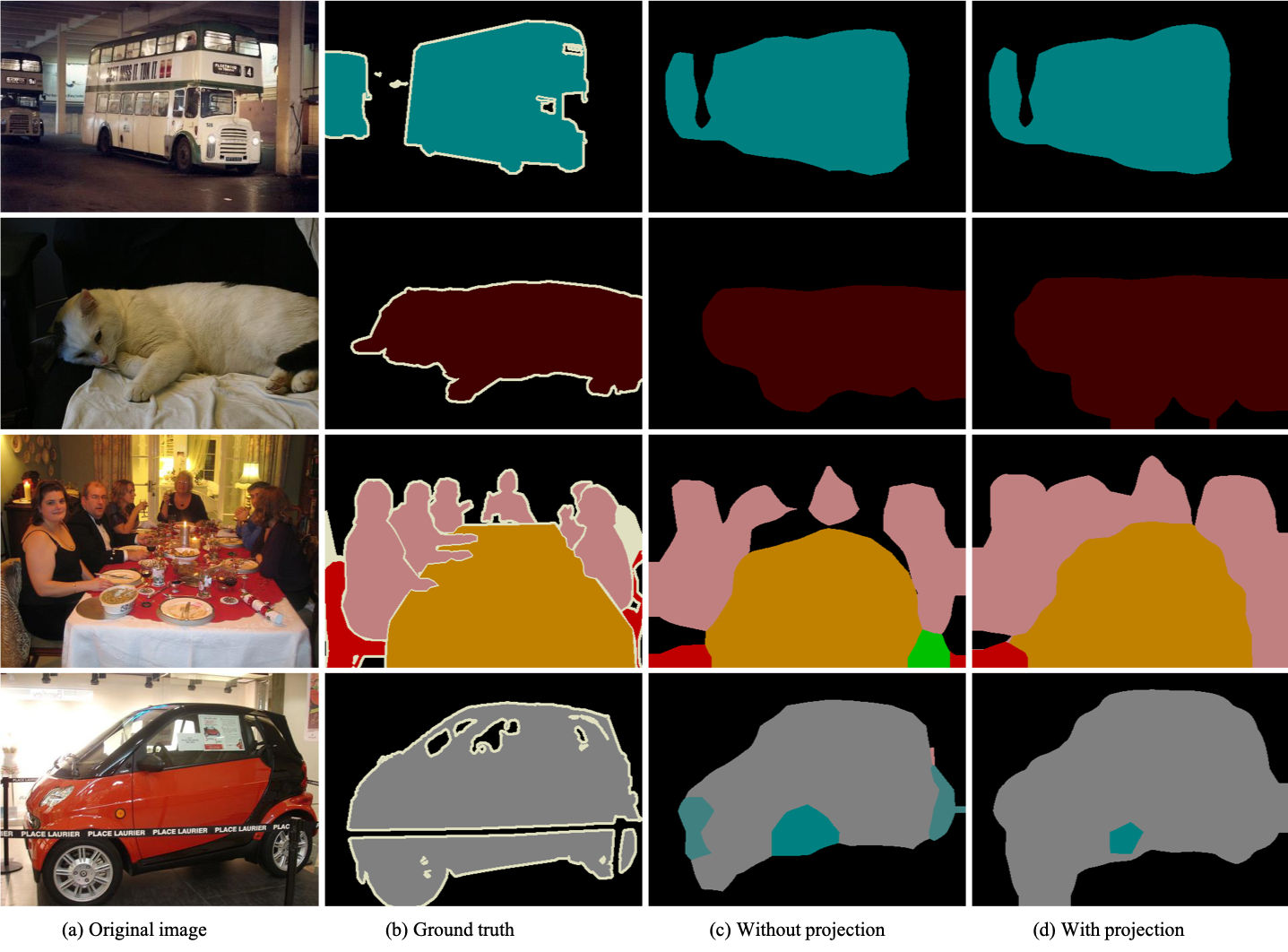}
    \caption{Qualitative results of experiments conducted using the VOC 2012 dataset. We show the original image, ground truth, our trained classifier without DP and DP layers. Note that the DP layers successfully revise the objects much better than those results without DP layers.}
    \label{fig:my_label}
\end{figure*}

Besides, the visualization of the qualitative results is shown in figure \ref{fig:my_label}. We notice that the proposed differentiable projection layer can suppress the pixels not listed in the image-level label without any parameter tuning. For example, the predicted segmentation of the VGG has some pixels labeled as a class (green pixels) not shown in the ground truth of the third image, and the results after adjusted by differentiable projection layer suppress the wrong label. Besides, in the fourth image, we notice that although the wrong label is not entirely suppressed, its size is smaller than the prediction of the VGG. Therefore, even without parameter tuning, the results are better than conventional VGG.
\begin{table*}[h]
    \centering
    \caption{Results on PASCAL VOC 2012 test. We compare our results to the fully supervised method and models with different DP layers.}
    \resizebox{\textwidth}{!}{
    \begin{tabular}{c c c c c c c c c c c c c c c c c c c c c c}
    \hline
    & mIoU & aero & bike & bird & boat & bottle & bus & car & cat & chair & cow & table & dog & horse & mbike & person & plant & sheep & sofa & train & tv \\
    \hline
    supervised & 48.07 & 34.84 & 6.07 & \textbf{45.32} & 36.10 & 26.94 & 70.28 & \textbf{44.80} & 58.44 & 11.41 & 54.90 & 43.91 & 49.62 & 39.22 & \textbf{53.01} & 39.74 & 11.25 & \textbf{46.13} & 44.08 & 44.59 & \textbf{40.35} \\
    \hline
    supervised (20 proj) & 47.73 & \textbf{35.05} & \textbf{6.74} & 42.14 & \textbf{37.05} & \textbf{28.44} & 69.37 & 44.40 & 51.15 & 11.23 & \textbf{60.97} & \textbf{46.59} & 49.97 & \textbf{40.83} & 50.49 & 37.71 & \textbf{12.38} & 44.44 & 44.32 & \textbf{47.35} & 39.06 \\
    \hline
    supervised(5 proj) & \textbf{49.20} & 34.50 & 6.26 & 44.25 & 36.77 & 27.18 & 70.53 & 44.26 & 56.61 & 11.69 & 55.22 & 44.21 & 49.75 & 39.57 & 52.38 & 39.55 & 11.67 & \textbf{45.74} & 44.39 & 45.55 & 39.49 \\
    \hline
    supervised (10 proj) & 48.02 & 34.43 & 6.40 & 43.61 & 36.65 & 27.61 & \textbf{70.90} & 44.44 & 55.23 & \textbf{11.87} & 56.93 & 45.18 & 50.00 & 40.10 & 52.05 & 39.26 & 11.88 & 45.52 & \textbf{44.55} & 46.33 & 39.37 \\
    \hline
    supervised (3 proj) & 47.87 & 34.59 & 6.20 & 44.49 & 36.86 & 27.01 & 70.23 & 44.17 & 57.07 & 11.63 & 54.69 & 43.95 & 49.61 & 39.40 & 52.54 & 39.61 & 11.56 & 45.74 & 44.31 & 45.31 & 39.52 \\
    \hline
    proj tuned(3 1e-6) & 47.84 & 34.21 & 6.06 & 44.61 & 36.71 & 26.94 & 70.17 & 44.22 & 57.26 & 11.47 & 54.63 & 43.78 & 49.68 & 39.37 & 52.56 & 39.71 & 11.37 & 45.71 & 44.28 & 45.09 & 39.31 \\
    \hline
    proj tuned(3 1e-5) & 46.90 & 29.76 & 3.92 & 45.28 & 34.31 & 25.94 & 69.33 & 43.73 & \textbf{59.11} & 10.22 & 52.90 & 42.55 & \textbf{50.15} & 38.52 & 51.28 & \textbf{40.16} & 10.14 & 44.99 & 43.03 & 41.71 & 37.29 \\
    \hline
    \end{tabular}}
    \label{tab:my_label}
\end{table*}

\section{Conclusion}
This paper aims to solve problems with linear constraints as a kind of prior knowledge in a more generalized form. We discuss the necessity of solving this kind of constrained problem under KKT conditions widely used before.  Also, we show that in linear constrained cases, we could use a projection method to efficiently build a differentiable projection layer of the DNN. We use the dot product of errors of both the actual output of the DNN and projected output to train the DNN. Besides, when the actual output is in the feasible region, we show that conventional DNN is a particular case of this projection DNN. Then, we conduct numerical experiments using a randomly generated synthetical dataset to evaluate the performance of the PDNN, including a comparison with a convention DNN and some different simple modifications. The experimental results show the effectiveness and robustness of the PDNN. Then, we also conduct image segmentation experiments to show the utility of the proposed DPDNN model. Currently, the projection layer proposed is differentiable, while there are only fixed parameters. In the future, we will investigate the improvement of this projection layer to achieve a projection layer structure that consists of learnable parameters. Also, we will evaluate our methods for some real-world applications.


%



\ifCLASSOPTIONcaptionsoff
  \newpage
\fi



%


%

\label{Bibliography}


\bibliographystyle{IEEEtran} 

\bibliography{main.bbl}
\end{document}